# A Multi-Modal Deep Learning Framework for Colorectal Pathology Diagnosis: Integrating Histological and Colonoscopy Data in a Pilot Study

Krithik Ramesh* and Ritvik Koneru*
*University of North Texas
Denton, United States of America
{KrithikRamesh, RitvikKoneru}@my.unt.edu

*Abstract*—Colorectal diseases, including inflammatory conditions and neoplasms, require quick, accurate care to be effectively treated. Traditional diagnostic pipelines require extensive preparation and rely on separate, individual evaluations on histological images and colonoscopy footage, introducing possible variability and inefficiencies. This pilot study proposes a unified deep learning network that uses convolutional neural networks (CNNs) to classify both histopathological slides and colonoscopy video frames in one pipeline. The pipeline integrates class-balancing learning, robust augmentation, and calibration methods to ensure accurate results. Static colon histology images were taken from the PathMNIST dataset, and the lower gastrointestinal (colonoscopy) videos were drawn from the HyperKvasir dataset. The CNN architecture used was ResNet-50. This study demonstrates an interpretable and reproducible diagnostic pipeline that unifies multiple diagnostic modalities to advance and ease the detection of colorectal diseases.

I. INTRODUCTION

Colorectal diseases are one of the leading causes of morbidity and mortality worldwide. Colorectal cancer was ranked third in most commonly diagnosed cancers and second in cancer-related deaths based on a study done in 2021 [1]. Therefore, diagnosing colorectal diseases is crucial, particularly for precancerous polyps, inflammatory ulcerative colitis, and other related malignant conditions. Polyps are malignant tissue growths on the inner lining of the colon and rectum. They may progress to colorectal cancer if not detected and treated early. Polyps can vary in morphology and pathology, making them hard to detect without endoscopic screening. Colitis refers to the inflammation of the colon lining and includes a spectrum of diseases. This spectrum includes ulcerative colitis and microscopic colitis. Although not a pathway to colorectal cancer, colitis can cause symptoms ranging from mild discomfort to severe abdominal pain, and, in severe cases, rectal bleeding. Therefore, the quick diagnosis of these diseases is crucial for the appropriate treatment and long-term health of patients with these conditions.

Traditionally, diagnostic workflows rely on manual interpretation of histological biopsies and colonoscopy videos. But, these methods tend to be resource-intensive, require specialists, and can be impacted by observer variabilities [2]. The misdiagnosis or misinterpretation of these conditions can lead to the progression of otherwise treatable conditions.

Artificial Intelligence (AI), specifically deep learning methods such as convolutional neural networks (CNNs), has shown promise in diagnosing and classifying medical imaging [3]. They have shown to enhance and automate the diagnostic accuracy. Although CNNs have demonstrated high accuracy in the segmentation and classification of image-based tasks like skin lesion and tumor detection, their application in multi-modal workflows like combining dynamic video data with static histopathology remains underdeveloped [4].

In this pilot study, the goal is to design and evaluate a unified, multi-modal deep learning pipeline that classifies colorectal pathology from both histological slides and colonoscopy video frames. Through the use of the PathMNIST dataset and the HyperKvasir dataset, the pipeline was able to train and output preliminary results that can be learned from [5], [6]. This study identified the weaknesses and strengths of the current pipeline, which will be improved on in the confirmatory study.

II. METHODOLOGY

*A. Project Initialization and Dataset Overview*

The project was initialized within Google Colab, an easily accessible and free-to-use Jupyter Notebook service that makes it easy to replicate studies. First, all of the dependencies were installed to make sure all of the packages were ready to use on the platform. The next step was to link the Google Colab notebook to Google Drive, so models and results could be securely saved.

The core static colon images used were from the public dataset PathMNIST, which is part of the MedMNIST v2 collection. PathMNIST consists of 107,180 hematoxylin and eosin (H&E) stained red, green, and blue (RGB) image patches from colorectal whole-slide images (WSIs) downsized to 28x28 pixels. PathMNIST had each sample labelled as one of nine tissue classes representing diverse categories: background, mucus, smooth muscle, epithelium, immune cells, debris, connective tissue, adipose, and cancerous tissues. These labels come from pixel-level annotations that are mapped to patch-level tiles, preserving coherence in the histological structures. The dataset was loaded into the notebook via the medmnist PyPI package. This handles all of the preprocessing and standardization of the data.

The lower gastrointestinal videos were extracted from HyperKvasir, a large public dataset containing gastrointestinal images and videos. For this study, the focus was on the lower gastrointestinal (colonoscopic) videos, focusing on two diagnostic categories: polyp and colitis detection. Frames were sampled at a rate of one frame per second from 74 polyp videos and 11 colitis videos extracted from HyperKvasir. These frames were then stored in Google Drive for easy access from Google Colab. The final dataset contained 6,100 training frames, 760 validation frames, and 1,040 testing frames. All frames were then resized to fit the 224x224 pixel input that ResNet-50 required [7]. This subset of HyperKvasir reflects the motion blur, low-light conditions, and inter-frame variability present in real-world colonoscopy footage. The use of this data helped mimic and evaluate performance on real-world colonoscopy footage.

*B. Preprocessing and Augmentation*

Preprocessing data ensures that the model inputs are accurate and numerically stable. The colonoscopy images from PathMNIST were resized from their original 28x28 pixel format to the 224x224 pixel format via bilinear interpolation to fit the default input resolution of the ImageNet-trained, ImageNet is an open source large collection of natural images, ResNet-50 model used in this study [8]. Since each histopathology image is an RGB image, it contains three 2D arrays (red, green, and blue). Each of these channels stores intensity values between 0 and 255, so, with this knowledge, by calculating the empirical mean and standard deviation of each of these channels across all of the training images in PathMNIST, each channel is centered and scaled (normalization), which helps training converge quickly and more reliably. This is an example of the equation being used for a red channel:

$$R_{\text{normalized}} = \frac{R_{\text{original}} - \mu_R}{\sigma_R}$$

(1)

$R_{\text{normalized}}$: the final value used as an input for ResNet-50.

$R_{\text{original}}$: the original red channel pixel value at a given position

$\mu_R$: the empirical mean of the red channel

$\sigma_R$: the standard deviation of the red channel

Although the HyperKvasir frames were originally at higher resolution, they were resized down to 224x224 via the same method as the PathMNIST images. Normalization for the frames followed ImageNet protocol: mean=[0.485, 0.456, 0.406], std=[0.229, 0.224, 0.225]. This facilitated smooth transfer learning from the pre-trained weights from ImageNet.

After this preprocessing, stochastic augmentation was implemented to enhance the generalization and mitigate the overfitting of ResNet-50. The augmentation pipeline included random horizontal flipping, random rotations, and color jittering. The augmentations were added exclusively to the training data through Torch Vision custom Python code. The validation and test sets remained unaltered to ensure the performance metrics reflected the true generalization of the model. Specifically for the frames derived from HyperKvasir, augmentation intensity was constrained due to the inherent variability of the real-world video-derived frames.

*C. Model Architecture and Training*

The core model, previously mentioned, used in both modalities was the ResNet-50 CNN architecture that was pre-trained on ImageNet. The original ImageNet fully connected (FC) layer was replaced by a new FC layer, matching the new output dimensions needed for each modality: 9 classes for PathMNIST and 2 for

HyperKvasir. ResNet-50 was the model used in this study due to its proven success and transferability in both natural and medical image classifications.

After the use of transfer learning was employed by initializing the CNN's backbone with ImageNet weights, fine-tuning was performed on the datasets using backpropagation. All of the training was conducted on the base, free CPU offered by Google Colab in order to maximize reproducibility.

Both of the models were trained using the Adam optimizer with an initial learning rate of 1e-4 and a weight decay of 1e-4. This learning rate was adjusted using a ReduceLROnPlateau scheduler, which decayed the learning rate by a factor of 0.5 if the validation loss did not improve after 1 epoch. The batch size of training examples processed together in a forward and backward pass in the CNN was 128 for images for PathMNIST and 64 frames for HyperKvasir. To ensure true generalization on the HyperKvasir trained ResNet-50 model, five random training seeds (42, 52, 62, 72, and 82) were implemented during training. Each of the seeds had controlled weight initializations and data shuffling, which promoted an estimate of variance across distinct runs. Early stopping was employed in the training of both models if stagnant validation accuracy was detected: 3 epochs for PathMNIST data and 5 epochs for HyperKvasir data. The maximum number of epochs was set to 20 for PathMNIST and 50 for HyperKvasir. Model checkpoints were saved to Google Drive based on the best validation loss numbers.

### D. Calibration and Explainability

To improve the reliability of the predictions on the PathMNIST data, a post-hoc temperature scaling was applied after training. A scalar temperature parameter ($T$) was optimized on the validation set to minimize negative log-likelihood. This is represented when logits ($z$) pass through a softmax function scaled by $T$:

$$P(y\,|\,x) = \text{softmax}\left(\frac{z}{T}\right)$$

(2)

After training, validation, and testing on the HyperKvasir data, a technique called Gradient-weighted Class Activation Mapping (Grad-CAM) was implemented to show the model's explainability [11].

### III. RESULTS AND DISCUSSION

#### A. Quantitative Performance Analysis

Both modalities achieved consistent results and accuracies. The ResNet-50 trained on PathMNIST data achieved a final, after calibration, test accuracy (7,180 samples) of 93.68%, a macro Area Under the Receiver Operating Characteristic Curve (AUC) of 0.9958, a macro F1-score of 0.9083, and a weighted one-vs-rest AUC of 0.9958. The highest validation accuracy and loss were achieved while training the model: Accuracy was 99.01%, and loss was 0.0289.

The training was stopped at epoch 10, showing rapid early improvement. This confirms excellent segmentation capabilities across multiple imbalanced classes. Before applying temperature calibration, the expected calibration error (ECE) was 0.057, but it improved to 0.030 after, improving alignment between the confidence and accuracy of the model.

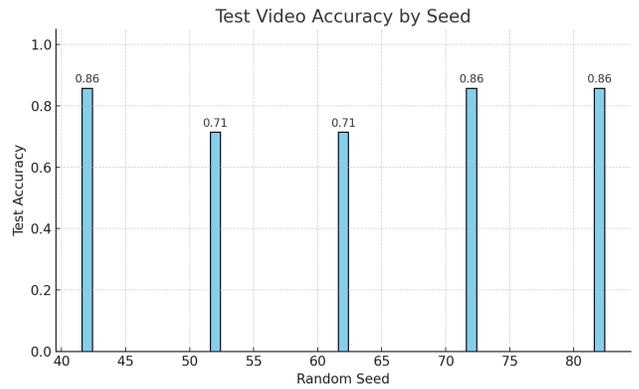

Fig. 1. HyperKvasir Test-Video Accuracies

The model trained on the HyperKvasir dataset yielded test-video accuracies of 85.7%, 71.4%, 71.4%, 85.7%, and 85.7% across all five random seeds (refer to Fig. 1). To calculate this, the model first predicted the class of each frame in each clip. Then, those predictions were aggregated into a single prediction per video. The prediction per video was then compared to the actual classification of the video to determine the test-video accuracy. Since there were 7 test videos, this means that 5-6 videos were classified correctly in each seed. The training dynamics across the seeds consistently showed rapid initial improvement, and early stopping was triggered at epoch 13. Several seeds achieved a validation accuracy of 1.00 before stabilizing while training. These results indicate stable generalization despite the limited amount of videos to train, validate, and test on. Frame-level confidence analysis across eight videos showed correct classification in 5 of 8 cases. At times, misclassified polyps were classified as colitis at a moderate confidence level (0.5-0.9). Correct predictions typically yielded higher confidence scores (>0.90), showing calibration robustness.

*B. Explainability with Grad-CAM*

Grad-CAM overlays (refer to Fig. 2-5) showed that the model focused on mucosal disruptions, raised lesions, and vascular patterning associated with inflammation. At times, the Grad-CAM was confused by foreign objects like tubes being inserted into the colon, causing issues with classification.

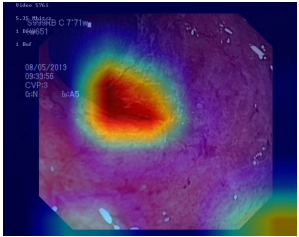

Fig. 2. Grad-CAM of colitis frame

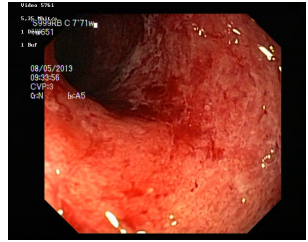

Fig. 3. Colitis frame

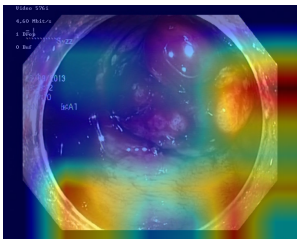

Fig. 4. Confused Grad-CAM of polyps frame

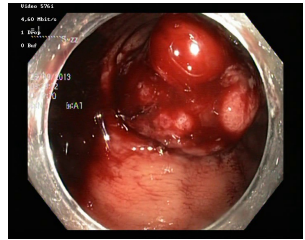

Fig. 5. Polyps frame

*C. Limitations and Future Work*

This study is inherently limited due to its pilot nature. The HyperKvasir data subset used in this study was small and lacked pixel-level annotations. Furthermore, while PathMNIST offers standardization, its resolution and format may reduce transferability to WSIs. While previous work has explored deep learning for colorectal cancer detection from colonoscopy and histopathological frames on a different dataset and a ResNet model, this pipeline only focuses on colitis and polyps specifically, while using a more accessible and reproducible pipeline [9]. Based on [9], this study's test accuracy (93.68%) is higher than theirs (92%), showing an improved pipeline.

Future work will involve scaling the video classification with more colonoscopy videos with less noise and incorporating temporal modeling architectures. First train on frames with minimal noise, then incorporate videos with noise to properly make sure ResNet-50 trains on the real data first before being exposed to data with real-world noise. For the histological images, moving more toward slide-level diagnosis would increase and enhance clinical relevance. Integrating uncertainty quantifications into both modalities will help quantify the confidence of the predictions and results. Finally, the end goal is to build a graphical user interface (GUI) that is easily accessible to clinicians and researchers (refer to Fig. 6 and Fig. 7).

## IV. CONCLUSION

In this pilot study, a multi-modal diagnostic model skilled at classifying both histological images and frames from colonoscopy videos was developed. The model backbone used was ResNet-50. The results of the research confirm strong performance on PathMNIST (with a test accuracy of 93.68% and macro AUC 0.9958) and good performance, given the data constraints, on HyperKvasir (with an average video accuracy of 79.98% across varying seeds). This shows that pretrained architectures (ResNet-50 on ImageNet) have the capability to generalize across WSI patches and video footage in colorectal diagnosis. Furthermore, post-hoc calibration markedly improved model reliability, and Grad-CAM heatmaps provided interpretability through clinically meaningful activation patterns.

While current findings are promising, the experimental parameters of this initial study are limited to sample size, video quality, and image resolution. This research study highlights a future framework for designing robust and explainable computer vision systems that can potentially enhance clinical workflows for the diagnosis and treatment of colorectal diseases.

Future studies will aim to improve this system by incorporating detailed video documentation with minimal noise and real-time inferential abilities. With the incorporation of cross-modal diagnostic features, this pipeline gives the groundwork for sophisticated and improved medical pipelines that take into account patient-to-patient variability and clinical risk factors.